\documentclass[runningheads]{llncs}
\usepackage{graphicx}
\usepackage{subcaption}
\usepackage{float}
\usepackage{rotating}
\usepackage{color,soul}
\usepackage{lscape}
\usepackage{tikz}
\usepackage{amsmath}
\usepackage{amsfonts}
\usepackage{multirow}
\usepackage{fancyhdr}
\usepackage[lined,boxed]{algorithm2e}

\begin{document}
\title{Towards Symbolic Time Series Representation Improved by Kernel Density Estimators}
%
%
\author{Matej Kloska\inst{1} \and 
Viera Rozinajova\inst{2}}
\authorrunning{Matej Kloska and Viera Rozinajova}
%
\institute{
    Faculty of Informatics and Information Technologies, \\
    Slovak University of Technology in Bratislava, \\
    Ilkovičova 2, 842 16 Bratislava, Slovakia\\
    \email{matej.kloska@stuba.sk} \\
    \url{https://www.fiit.stuba.sk} \and
    Kempelen Institute of Intelligent Technologies, \\
    Mlynské Nivy II.18890/5, \\
    811 09 Bratislava, Slovakia \\
    \email{viera.rozinajova@kinit.sk} \\
    \url{https://www.kinit.sk}
}
\maketitle  
\begin{abstract}
    This paper deals with symbolic time series representation. It builds up on the popular mapping technique Symbolic Aggregate approXimation algorithm (SAX), which is extensively utilized in sequence classification, pattern mining, anomaly detection, time series indexing and other data mining tasks. However, the disadvantage of this method is, that it works reliably only for time series with Gaussian-like distribution.
    In our previous work (Kloska and Rozinajova, dwSAX, 2020) we have proposed an improvement of SAX, called dwSAX, which can deal with Gaussian as well as non-Gaussian data distribution. Recently we have made further progress in our solution - edwSAX. Our goal was to optimally cover the information space by means of sufficient alphabet utilization; and to satisfy lower bounding criterion as tight as possible. We describe here our approach, including evaluation on commonly employed tasks such as time series reconstruction error and Euclidean distance lower bounding with promising improvements over SAX.

\end{abstract}

\thispagestyle{fancy}
\fancyhf{}
\headheight 40pt \headsep 10pt
\renewcommand{\headrulewidth}{0pt}
\lhead{\scriptsize This article has been published in a revised form in \emph{Transactions on Large-Scale Data- and Knowledge-Centered Systems L} [http://doi.org/10.1007/978-3-662-64553-6\_2]. This version is published under a Creative Commons CC-BY-NC-ND. No commercial re-distribution or re-use allowed. Derivative works cannot be distributed. © Matej Kloska, Viera Rozinajova.}

\section{Introduction}
It is generally known that huge amounts of data are generated on a daily basis, whereas streaming data make up a large part of them.  There are many fields, such as healthcare, finance, security, energy and industry, where intelligent analysis and data mining tasks play an important role. Frequently the data capture event observations taken according to the order of time - we usually talk about time series. As we often deal with continuous data stream, it is clear, that its representation poses a significant problem when processing them.

At the same time, time series usually capture feature rich, highly dimensional data which make processing tasks even harder. In our work, the high dimensionality is related to a high number of varying data points in time series often capturing time series over a long period of time covering different behavior trends such as different power consumption during day and night, lockdown periods or seasonal differences. All these trends contribute to highly dimensional time series --- which re difficult to process while capturing all significant properties over time. Dimensionality reduction and descriptive forms of time series representation are recognized as a possible solution for highly performing data mining tasks \cite{ref_koegh_dimensionality_reduction_2001}. A challenging area in the field of effective time series processing is their compact data presentation without sacrificing any significant information \cite{ref_wang_time_series_representation_survey_2013}. Symbolic representation of time series appears to be the solution to this problem giving us the opportunity to exploit longer time series periods without significantly higher computational resources for most of data mining tasks.

The Symbolic Aggregate approXimation algorithm (SAX) \cite{ref_sax_original} is one of the most popular symbolic mapping techniques for time series. SAX as a powerful unsupervised symbolic mapping technique is widely used due to its data adaptability. It is extensively utilized in sequence classification \cite{ref_senin_classification_2013}, pattern mining \cite{ref_fournier_viger_pattern_mining_2017}, anomaly detection \cite{ref_keogh_hotsax_2005} and many other data mining tasks \cite{ref_lin_experiencing_sax_2007,ref_shieh_indexable_sax,ref_tamura_2017}. However, this approach heavily relies on assumption that processed time series have Gaussian-like distribution \cite{ref_sax_original}. When time series distribution is non-Gaussian or skews over time, this method does not provide sufficient symbolic representation. 

In our previous work \cite{kloska2020distribution} we have proposed an improvement of SAX, called dwSAX, which can deal with Gaussian as well as non-Gaussian data distribution. Later, the similar approach was presented by Bountrogiannis et. al \cite{bountrogiannis2021data}. In this paper we introduce an improved technique for symbol breakpoints and centroids selection which contributes to more efficient alphabet symbols utilization. The goal is to optimally cover the information space and prove that our extension satisfies lower bounding criterion for wide exploitation of our method. The method was evaluated on commonly employed tasks such as time series reconstruction and Euclidean distance lower bounding with promising improvements over SAX.

This paper is organized as follows. Section 2 describes original SAX method and techniques for data distribution estimation. Section 3 introduces dwSAX - our extension to SAX method and its further improvement, which represents the core of this paper --- edwSAX. Section 4 contains an experimental evaluation of the proposed method on time series reconstruction error and tightness of lower bound tasks compared to the original SAX method. Finally, Section 5 offers some conclusions and suggestions for future work.

\section{Related work}
    One of efficient data stream processing problems is their high dimensionality, too high number of data points due to various reasons such as mixed observed behaviors or gradual change over time. Possible solution to this issue is efficient symbolic representation of  a high-dimensional data stream through a less dimensional symbolic data stream. In past decades, many different time series representations have been introduced. Lin et al. \cite{ref_sax_original} divided methods into data adaptive (eg. Piecewise Linear Approximation, Singular Value Decomposition, SAX) and non data adaptive (eg. Wavelets, Random Mappings, Discrete Fourier Transformation). Recent research \cite{ref_sato_pla_2017,ref_tamura_2017,ref_yang_haar_wavelet_2018,ref_eghan_haar_wavelet_2019,ref_mahmoudi_dft_2019} shows activities in both method families. In the following sections we discuss fundamentals of original SAX, and techniques for data distribution estimation.

\subsection{Symbolic representation - SAX}
\begin{figure}
        \centering
        \fbox{\includegraphics[width=1.0\textwidth]{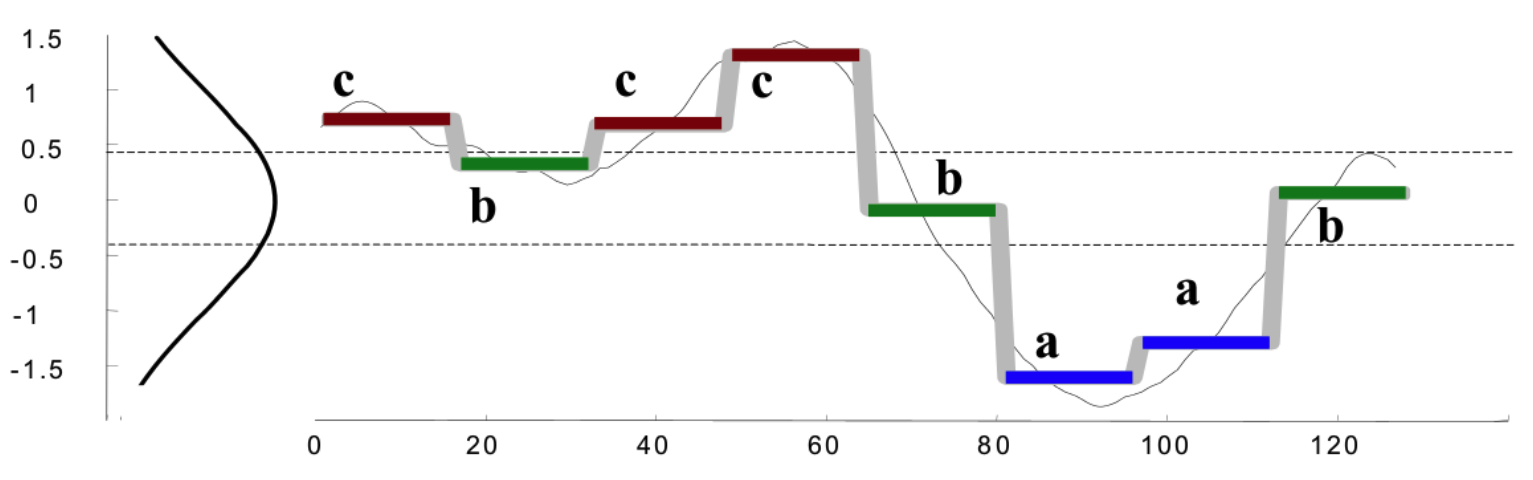}}
        \caption{Concept of Original SAX. Background grey thin line is replaced with bold line segments (PAA). PAA segments are mapped by normal distribution sketched on y axis into symbols, $a=(-\infty; -0.43\rangle,b=(-0.43; 0.43\rangle,c=(0.43, \infty)$ \cite{ref_sax_original}.}
        \label{fig:sax_concept}
    \end{figure}

    SAX is one of the best known algorithms for symbolic time series representation. This method makes it possible to represent any time series of length $n$ using a string of any length $w$ $(w \ll  n)$ with symbols from the predefined alphabet. Looking at the mentioned method, it consists of:
    \begin{enumerate}
        \item \textit{dimensionality reduction}: applying Piecewise Aggregate Approximation (PAA) \cite{ref_koegh_dimensionality_reduction_2001}, it significantly reduces dimensionality and preprocesses time series for further step;
        \item \textit{discretization}: mapping PAA segments into specific symbols from the alphabet based on a pre-computed mapping symbols table.
    \end{enumerate}
    Concept of this method is illustrated in Figure. \ref{fig:sax_concept}. Throughout this paper we use common notation used also in the original SAX paper \cite{ref_sax_original} which you can find in Table \ref{tab_sax_notation}.
    
    \begin{table}
        \centering
        \caption{A summarization of common notation used in this paper and the original SAX paper. \cite{ref_sax_original}}
        \label{tab_sax_notation}
        \begingroup
            \setlength{\tabcolsep}{7pt} 
            \renewcommand{\arraystretch}{1.5}
            \begin{tabular}{|l|l|}
            \hline
            C       & A time series $C = c_1, ..., c_n$ where $c_i \in \mathbb{R}$                                                                 \\ \hline
            $\bar{C}$ & \begin{tabular}[c]{@{}l@{}}A Piecewise Aggregate Approximation of a time series  $\bar{C} = \bar{c}_1, ...,\bar{c}_w$\end{tabular} \\ \hline
            $\hat{C}$ & \begin{tabular}[c]{@{}l@{}}A symbol representation of a time series $\hat{C}=\hat{c}_1, ...,\hat{c}_w$\end{tabular}                  \\ \hline
            \(w\)   & \begin{tabular}[c]{@{}l@{}}The number of PAA segments representing time series  C\end{tabular}                                     \\ \hline
            \(a\)   & Alphabet size (e.g., for the alphabet = \{a,b,c\}, \(a = 3\))                                                                       \\ \hline
            \end{tabular}
        \endgroup
    \end{table}

\subsubsection{Dimensionality reduction}
    Intuition based on the aforementioned description is to reduce time series from $n$ dimensions into $w$ dimensions. This goal is simply achieved by division of the time series into $w$ equal sized pieces - segments. For each piece, mean value is calculated, and this value represents the underlying vector of $w$ original values. Total vector of all pieces becomes a new reduced representation of the original time series.
    
    More formally, a time series $C$ of length $n$ can be reduced into a w-dimensional time series by a vector $\bar{C}=\bar{c}_1, ..., \bar{c}_w$ where $i^{th}$ element of $\bar{C}$ is calculated as follows \cite{ref_sax_original}:
    \begin{equation}
        \bar{c}_i=\frac{w}{n}\sum_{j=\frac{n}{w}(i-1)+1}^{\frac{n}{w}i}c_j
    \end{equation}
    
\subsubsection{Discretization}
    Discretization step replaces PAA segments obtained in the length reduction step by alphabet symbols. Assuming data are normalized before reduction (with zero mean) and have highly Gaussian-like distribution, the replacement is performed as follows. We first precompute a table of breakpoints. Lin et al. \cite{ref_sax_original} defined breakpoints as sorted  list  of numbers $B = \beta_1, ..., \beta_{a-1}$ such  that  the  area  under a $N(0,1)$ Gaussian curve from $\beta_i$ to $\beta_{i+1} = 1/a$ ($\beta_0$  and $\beta_a$  are defined as $-\infty$ and $\infty$, respectively).
    
    The Gaussian curve enables efficient breakpoints table precomputation, thus the discretization step is trivial in comparison to the single vector lookup operation.
    
    Finally, formal definition of SAX as proposed by Lin et al. \cite{ref_sax_original}: A subsequence $C$ of length $n$ can be represented as a word $\hat{C} = \hat{c}_i, ..., \hat{c}_w$ as follows. Let $\alpha_i$ denote the $i^{th}$ element of the alphabet, i.e., $\alpha_1 = a$ and $\alpha_2 = b$. Then the mapping from a PAA approximation $C$ to a word $\hat{C}$ is obtained as follows:
    \begin{equation}
        \hat{c}_i = \alpha_j,\hspace{10pt}iif\hspace{15pt}\beta_{j-1} \leq \bar{c}_i < \beta_j
    \end{equation}

\subsection{Techniques for distribution estimation}
    In the previous section we discussed internals of the original SAX method. SAX uses a Gaussian distribution to derive the regional breakpoints resulting in the generation of an equiprobable set of symbols. As we already mentioned, our method wants to make a new SAX method Gaussian distribution requirement free. In this section we want to mention other methods on how to estimate data distribution and, based on them, improve SAX by a different way of setting the regional breakpoints. At the beginning, we want to state a common intuition to graphically represent data distribution - histogram plotting from underlying time series data points.
    
\begin{figure}[H]
        \centering
        \fbox{\includegraphics[width=.47\textwidth]{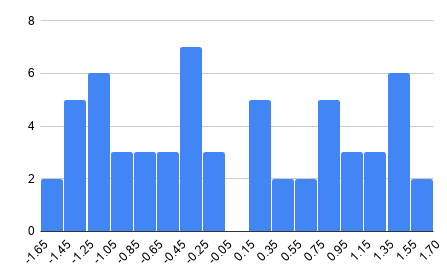}}\hfill
        \fbox{\includegraphics[width=.47\textwidth]{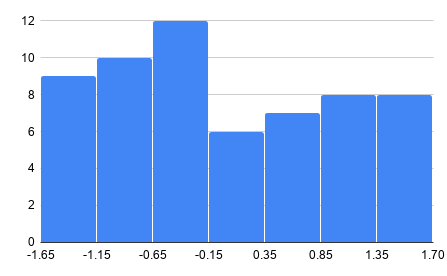}}
        
        \caption{Comparison of two histograms for the same dataset with bin widths 0.2 and 0.5 respectively. Incorrectly selected bin width causes visually different data distribution.}
        \label{fig:histogram_issue}
    \end{figure}

    The histogram is a former nonparametric density estimator with strong use in exploratory data analysis for displaying and summarizing data. 
    Bin width is an important parameter that needs selection prior histogram construction. It is evident that the choice of the bin width has a strong effect on the shape of the resulting histogram. The example for different bin width selection you can find in Figure \ref{fig:histogram_issue}.
    
    There are several ways to determine optimal bin width $\hat{h}$ with $n$ observed instances such as \cite{ref_wand_histogram_bin_width_1997}:
    \begin{equation}
        \hat{h}=\frac{range\_of\_data}{1 + log_2n}
    \end{equation}
    or alternatively more general formula based on Mean Integrated Squared Error (MISE):
    \begin{equation}
        \hat{h}=\hat{C}n^{-1/3},
    \end{equation}
    where $\hat{C}$ is any selected statistic. Most known example of above mentioned formula is normal reference rule \cite{ref_scott_optimal_histogram_1979,ref_wand_histogram_bin_width_1997}:
    \begin{equation}
        \hat{h}=3.49\hat{\sigma}n^{-1/3},
    \end{equation}
    where $\hat{\sigma}$ is an estimate of the standard deviation.
    
\begin{figure}[H]
        \centering
        \includegraphics[width=.85\textwidth]{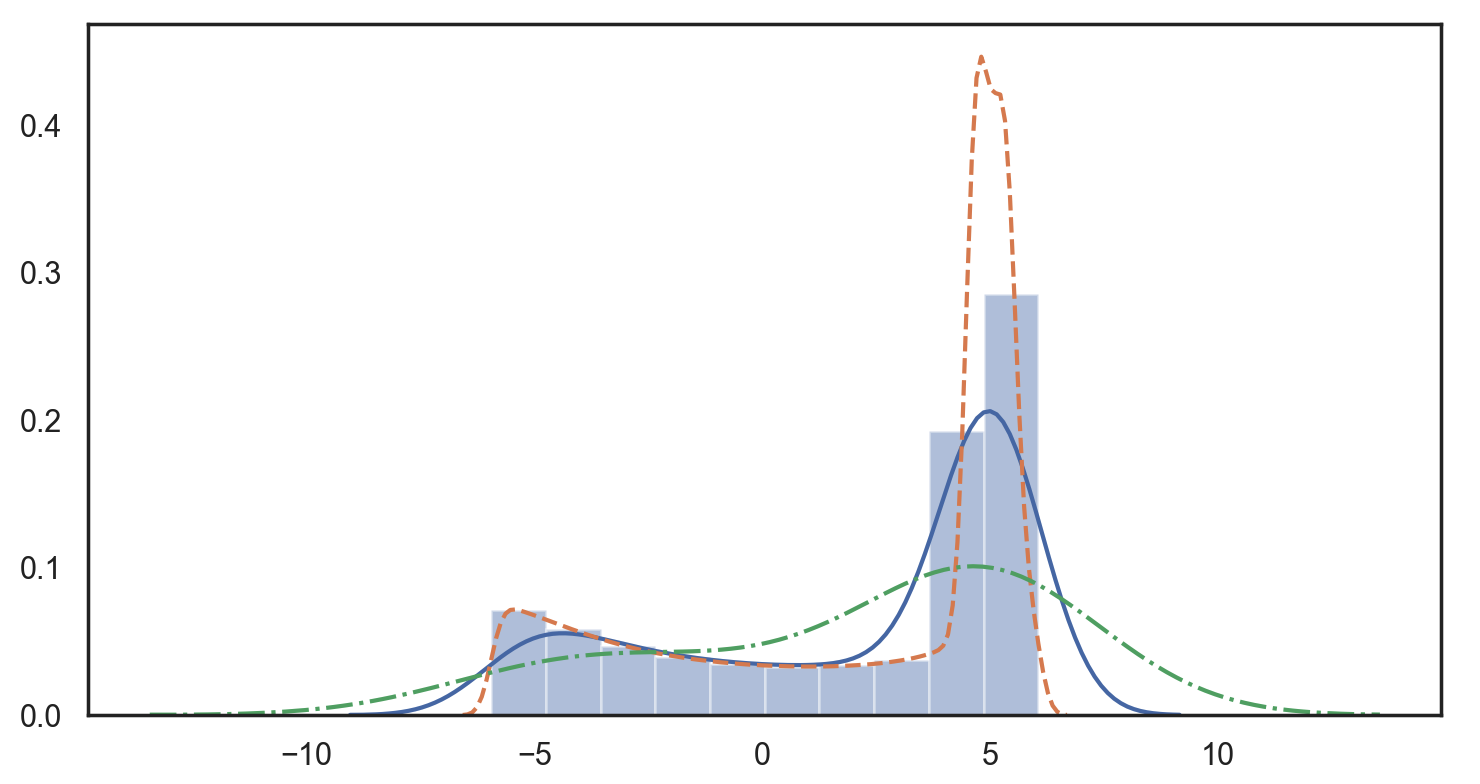}
        
        \caption{The implications of different KDE bandwidth parameter selection Gaussian kernel. Optimal (blue line), oversmoothing (green line) and undersmoothing (orange line) bandwidths.}
        \label{fig:kde_examples}
    \end{figure}

    In the past four decades there was research in the field of distribution estimation using continuous functions - density estimators.
    Intuition behind kernel estimators is to describe underlying data histogram with a smooth continuous line. More formally, given a set of $N$ training data ${y_n, n = 1, ..., N}$, a kernel density estimator (KDE), with the kernel function $K$ and a bandwidth parameter $h$, ($h \in R; h > 0$), gives the estimated density $\hat{f}(y)$ for data $y$ as follows \cite{ref_hwang_nonparametric_density_1994}:
    \begin{equation}
        \hat{f}(y) = \frac{1}{N}\sum_{n=1}^{N}K(\frac{y-y_n}{h})
    \end{equation}
    Kernel function $K$ should satisfy positivity and integrate-to-one constraints \cite{ref_hwang_nonparametric_density_1994}:
    \begin{equation}
        K(u) \geq 0, \hspace{15pt} \int_{R^+}K(u)du = 1
    \end{equation}
    The quality of a kernel estimate depends less on the chosen $K$ than on the bandwidth value $h$. It is crucial to choose the most suitable bandwidth because a value that is too small or too large will result in not useful estimation. Small values of $h$ lead to undersmoothing estimates while larger $h$ values lead to oversmoothing \cite{ref_jones_bandwidth_selection_survey_1996}. Figure \ref{fig:kde_examples}. illustrates possible cases of incorrectly selected bandwidth parameter $h$. To overcome this issue, a number of different bandwidth selection methods such as Silverman's rule of thumb \cite{silverman2018density}, Scott’s rule of thumb \cite{scott2010scott} or Improved Sheather-Jones bandwidth selection \cite{sheather1991reliable} have been proposed.
    
    Table \ref{tab_kernels_list}. provides a list of most frequently used kernels with their support ranges. Selection of different kernel does not significantly change the shape of the estimated function, but its computational complexity might introduce significant overhead. Support range might be a constraint towards deciding which kernel is suitable for a specific case due to long-tail $\mathbb{R}$ support in cases such as Normal or Laplace kernels.
    
    \begin{table}
        \centering
        \caption{A list of commonly used kernel functions with their function formula and support range. \cite{jones1990performance}}
        \label{tab_kernels_list}
        \begingroup
            \setlength{\tabcolsep}{6pt} 
            \renewcommand{\arraystretch}{2}
            \begin{tabular}{|l|l|l|}
            \hline
            \textbf{Kernel}       & \textbf{$K(u)$} & \textbf{Support} \\ \hline
            Uniform      & $\frac{1}{2\sqrt{3}}$                                      & $\|u\| \leq 3$ \\ \hline
            Triangular   & $\frac{(1-\frac{\left | u \right |}{\sqrt{6}})}{\sqrt{6}}$ & $\|u\| \leq 6$ \\ \hline
            Epanechnikov & $\frac{3(1-\frac{u^{2}}{5})}{4\sqrt{5}}$                   & $\|u\| \leq 5$ \\ \hline
            Biweight     & $\frac{15(1-\frac{u^{2}}{7})^{2}}{16\sqrt{7}}$             & $\|u\| \leq 7$ \\ \hline
            Cosine       & $\frac{\sqrt{\pi^2-8} cos(\frac{\sqrt{\pi^2-8}x}{2}))}{4}$ & $\|u\| \leq \frac{\pi}{\sqrt{\pi^2-8}}$ \\ \hline
            Normal       & $\frac{2\pi}{\sqrt(2)}exp(-\frac{1}{2}x^2))$               & $\mathbb{R}$ \\ \hline
            Laplace      & $\frac{exp(-\sqrt{2}|x|)}{\sqrt{2}}$                       & $\mathbb{R}$ \\ \hline
            \end{tabular}
        \endgroup
    \end{table}

\section{Distribution-wise SAX (edwSAX)}
    Formerly proposed SAX method is not well suited for time series with non-Gaussian distribution. If we apply Gaussian distribution lookup vector for breakpoints, we get non-optimal, still feasible, breakpoints. Our modified implementation focuses on elimination of this deficiency. Algorithm \ref{alg:dwsax} illustrates overall main flow of the function where input is sequence for symbolization, word length, alphabet size and bandwidth for kernel estimator result is sax representation of input sequence. In the next sections we will discuss specific internals of our method:
    \begin{enumerate}
        \item \textit{probability density function estimation}: given normalized time series, we need to estimate probability density function to proceed with more precise breakpoints;
        \item \textit{breakpoints and centroids vector calculation}: having probability density function, the task is to calculate equiprobable breakpoints covering the domain of $pdf$;
        \item \textit{distance measure definition}: to prove our efficiency and indexing capability, we also define a modified distance measure proving lower bound euclidean distance criterion.
    \end{enumerate}
    
    \begin{algorithm}[H]
        \label{alg:dwsax}
        \KwData{Sequence, WordLength, AlphaSize, Bandwidth}
        \KwResult{SAX Repr}
        PAA Sequence = \textit{ApplyPAA}(Sequence, WordLength)\;
        PDF Estimate = \textit{EstimatePDF}(Sequence)\;
        Breakpoints Vector = \textit{CalculateBreakpoints}(AlphaSize, PDF Estimate)\;
        SAX Repr = \textit{Map}(PAA Sequence, Breakpoints Vector)\;
        \caption{edwSAX main algorithm}
    \end{algorithm}
    
\begin{figure}[!h]
        \centering
        \includegraphics[width=1.0\textwidth]{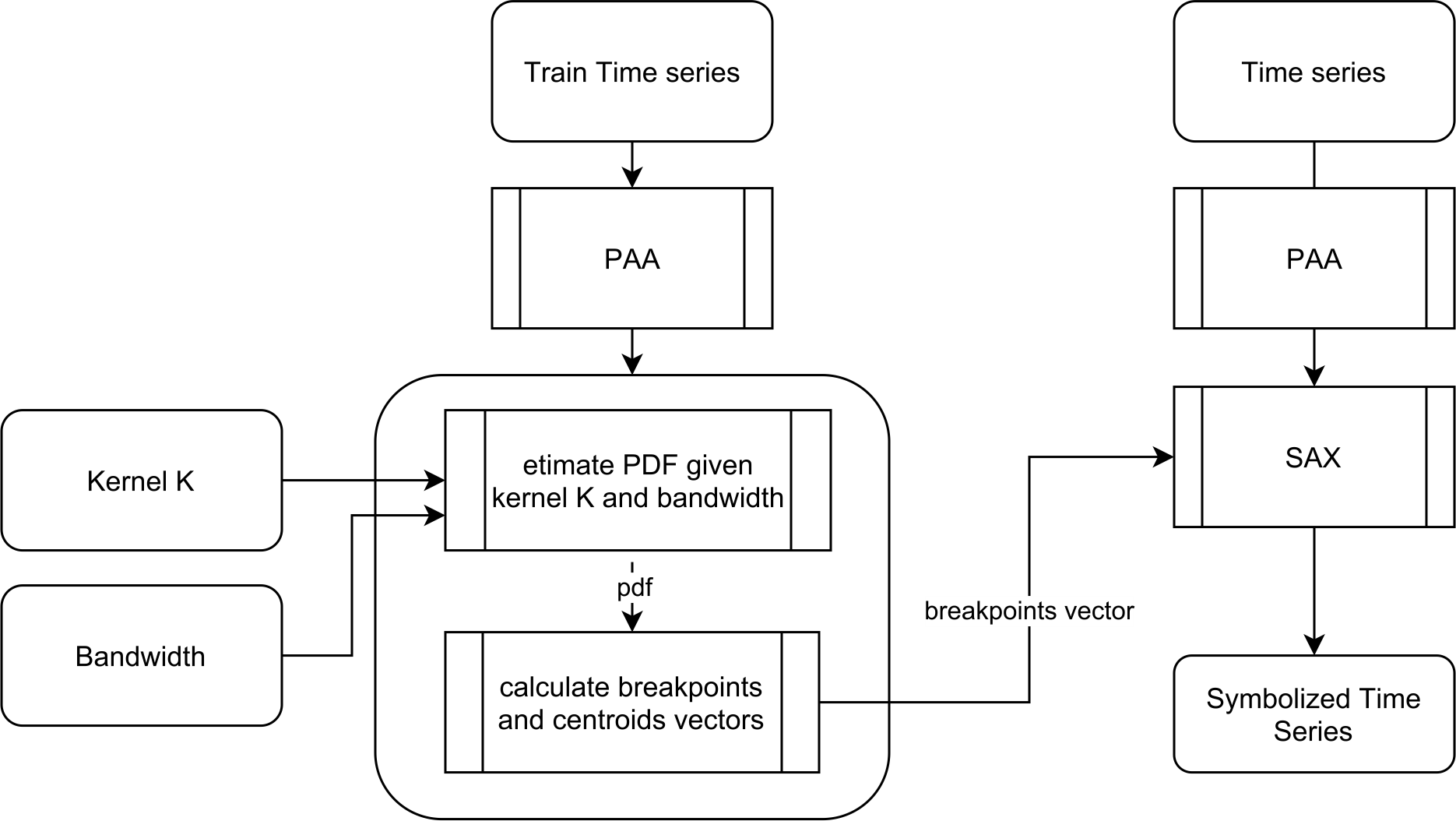}
            
        \caption{The flow of edwSAX method. In contrast to SAX, we need also a training phase (left top-down flow) to estimate $pdf$ and calculate the breakpoints vector. Once breakpoints vector is trained, we follow well-known SAX flow (right top-down float).}
        \label{fig:dwsax_flow}
    \end{figure}

    Our proposed method flow in contrast to SAX expects also training phase (Fig. \ref{fig:dwsax_flow}). At this phase, we need to estimate $pdf$ in order to correctly calculate breakpoints and centroids vectors. All further steps after training are identical with the former method. Keeping this fact in mind, we can virtually extend any other improved SAX method with our alternative breakpoints vector in order to be efficient on non-Gaussian-like data sets.
    
\subsection{Probability density estimation}
\begin{figure}[H]
        \centering
        \subcaptionbox{Example of SAX representation. \\ Resulting symbolic string: \textit{eeeeeeeeedbabaabbaab}}{%
            \fbox{\includegraphics[width=.75\textwidth]{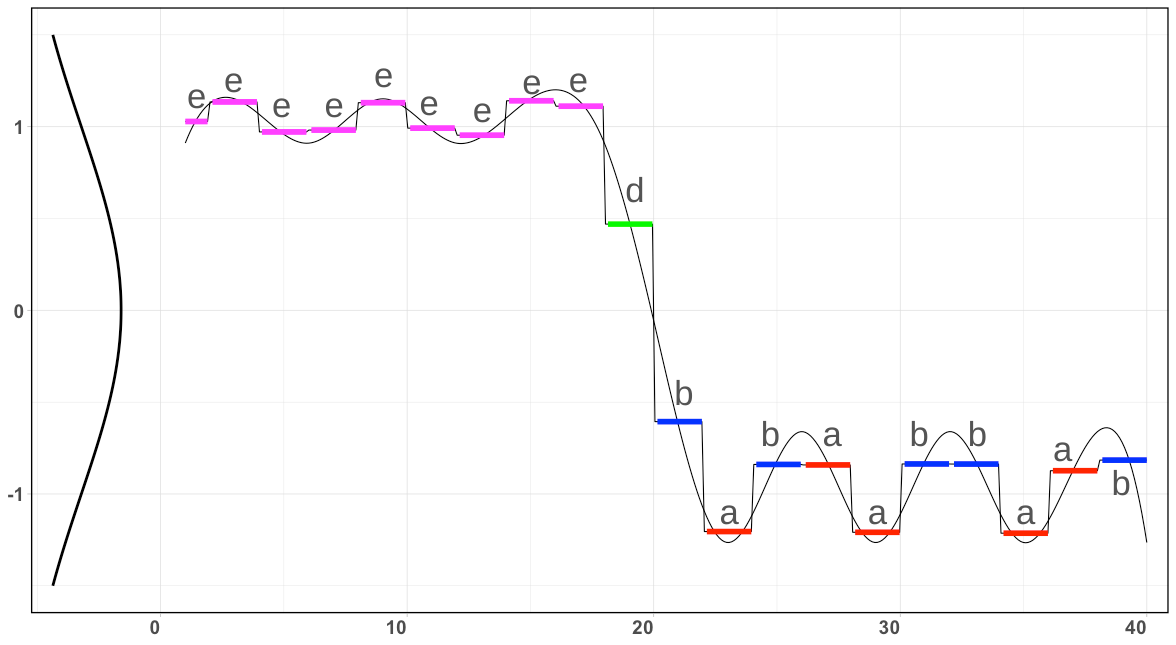}}%
        }\par\medskip
        \subcaptionbox{Example of edwSAX representation. \\ Resulting symbolic string: \textit{deddeddeecbabbabbabb}}{%
            \fbox{\includegraphics[width=.75\textwidth]{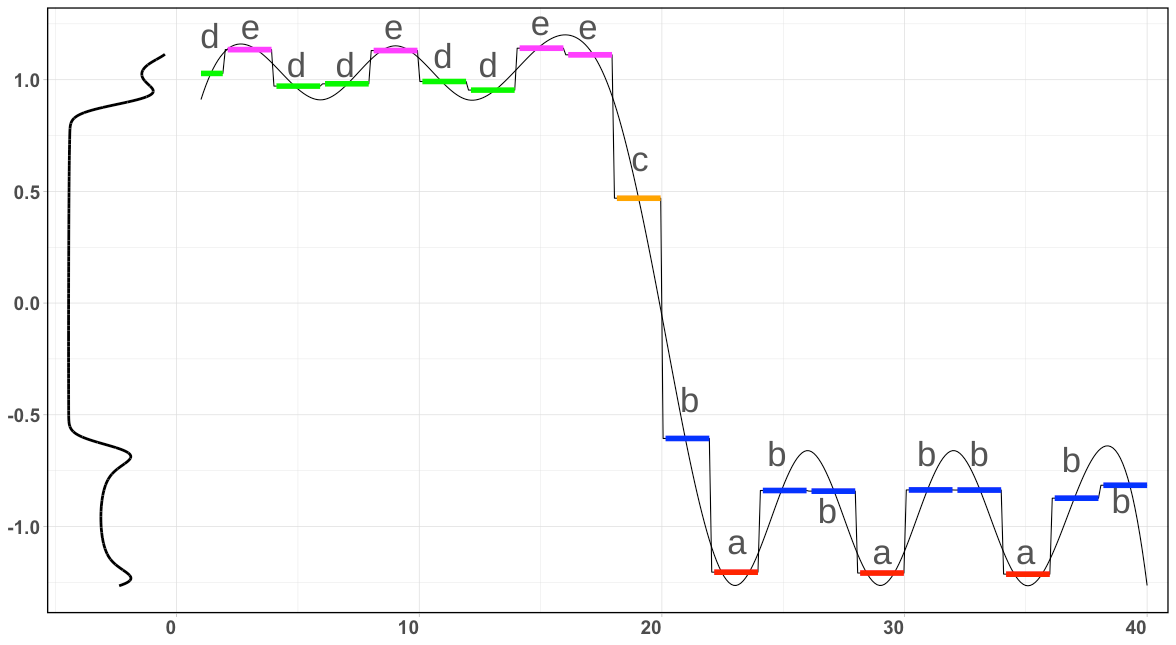}}%
        }
        
        \caption{Application of different $pdf$ on breakpoints selection. a) SAX with Gaussian distribution, b) edwSAX with KDE. KDE based breakpoints bring more precise symbolic representation compared with SAX with the same alphabet size 5 and PAA=2.}
        \label{fig:probability_sax_vs_dwsax}
    \end{figure}

    With transformed PAA time series we can proceed to probability density estimation. Having precisely estimated probability function will help us in further breakpoints vector calculation for the discretization procedure step. A naive solution to this problem seems to be histogram exploitation as its computational complexity is incomparably lower to the other methods such as KDE. The main drawback of histograms is their discrete representation which is not suitable for breakpoints interpolation. Our method for breakpoints calculation expects continuous probability function suitable for integral calculus with integrate-to-one constraint. KDE appears to be the solution to this problem. This method needs to specify a kernel function $K(\cdot)$ and a bandwidth parameter $h$. Selection of appropriate kernel function and bandwidth parameter depends on data and required precision of overall symbolic representation performance. To our best knowledge, Gaussian kernel gives the most relevant results and should be applied as the first possible option for KDE exploitation. In Table \ref{tab_kernels_list} we mentioned multiple options related to the most frequently used KDE kernels.
    However, for our implementation we advise to use Kernels with more compact support rather than Gaussian one. The default option is Epanechnikov kernel with Improved Sheather Jones (ISJ) index which outperforms asymptotically Silverman's rule of thumb on non-Gaussian-like data sets for bandwidth parameter selection. The difference between edwSAX and SAX itself is depicted in Figure \ref{fig:probability_sax_vs_dwsax}.
    
\subsection{Breakpoints and centroids vector calculation}
    Having estimated probability function using KDE, we can advance and estimate breakpoints based on probability distribution of time series with their respective representatives - centroids. The main goal is to efficiently compute those breakpoints as we do not apply only specific Gaussian distribution and its pre-computed values table. However, the idea for breakpoints selection is the same - select points from the KDE probability density function ($pdf$) such that they produce equal-sized areas under the KDE function curve. 
    
    \begin{definition}
        Let $a$ denote alphabet size, $pdf$ probability density function and $\beta_n$, $\beta_{n+1}$ any two consecutive breakpoints from breakpoints vector $B$. Then breakpoints vector $B$ is defined as a vector of ordered breakpoints $\beta$ such that $\beta_n, \beta_{n+1}$ follows:
        \begin{equation}
            \int_{\beta_n}^{\beta_{n+1}}pdf(y)dy = \frac{1}{a}
        \end{equation}
    \end{definition}
    
    With the computed breakpoints vector, we proceed in most suitable representatives - centroids calculation for each consecutive non-overlapping breakpoints pair. Most suitable representative represents in our case centroid $\gamma$, $\gamma \in \left \langle \beta_n; \beta_{n+1} \right \rangle$), fulfilling equiprobable condition for ranges $\left \langle \beta_n; \gamma \right \rangle$ and $\left \langle \gamma; \beta_{n+1} \right \rangle$:
    \begin{equation}
        \int_{\beta_n}^{\gamma}pdf(y)dy=\int_{\gamma}^{\beta_{n+1}}pdf(y)dy \hspace{10pt}iif\hspace{15pt} \gamma \in \left \langle \beta_n; \beta_{n+1} \right \rangle
        \label{eq:centroid_equal_probability}
    \end{equation}
    
    Computed centroids vector is stored and used along breakpoints vector during the reconstruction process, giving us lower reconstruction error by means of any well-known error metrics such as MAE or MSE. Breakpoints vector is also necessary in order to correctly calculate distance measure, defined in section \ref{sect:distance_measure}.
    
    Discretization process follows the same algorithm as proposed in the original SAX method. For reference see algorithm \ref{alg:dwsax_mapping}.

    \begin{algorithm}[H]
        \SetAlgoLined
        \label{alg:dwsax_mapping}
        \KwData{PAA Sequence, Breakpoints Vector B}
        \KwResult{SAX Representation}
        \ForEach{Segment in PAA Sequence} {
            \For{$i \leftarrow 2$ \KwTo Length(B) } {
                \If{ $B[i-1] \leq Segment$ \textbf{and} $Segment < B[i]$ }{
                    Append(SAX Representation, Alphabet[i])
                }
            }
        }
        \caption{edwSAX mapping procedure}
    \end{algorithm}

\subsection{Distance measure}
    \label{sect:distance_measure}
    Having defined the symbolic time series representation, we now define the similarity measure on the transformed data and we prove it lower bounds the Euclidean distance on the original data. Distance measure for edwSAX time series representation is based on the existing proved MINDIST distance \cite{ref_sax_original} - Euclidean distance adaption:
    \begin{equation}
        \label{eq:mindist_distance}
        MINDIST(\tilde{Q}, \tilde{C}) \equiv \sqrt{\frac{n}{w}}\sqrt{\sum_{i=1}^{w}(dist(\tilde{q_i}, \tilde{c_i}))^2}
    \end{equation}
    where $\tilde{Q}$, $\tilde{C}$ are symbolized time series, $n$ is original time series length, $w$ is word length (based on PAA parameter) and $dist(\tilde{q_i}, \tilde{c_i})$ is function returning minimal distance between symbols $\tilde{q_i}$ and $\tilde{c_i}$.
    
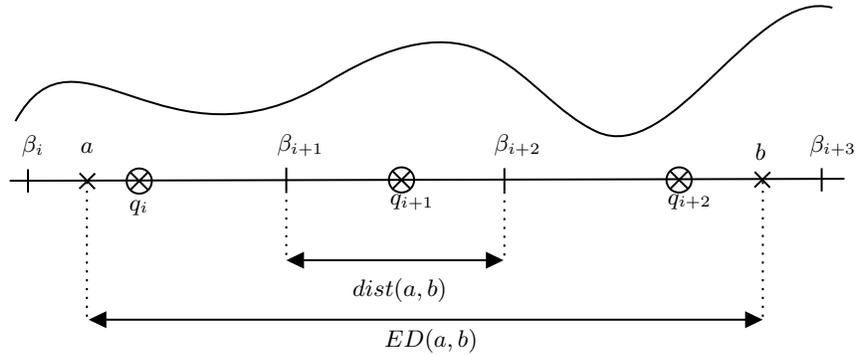
\begin{figure}[H]
        \centering

        \tikzset{every picture/.style={line width=0.75pt}} 
        
        \begin{tikzpicture}[x=0.75pt,y=0.75pt,yscale=-1,xscale=1]
            
            \draw    (110.6,140.6) -- (531.6,139.6) ;
            \draw    (119.83,134.79) -- (119.83,146.54) ;
            \draw    (250.11,133.94) -- (250.11,145.69) ;
            \draw    (520.11,134.22) -- (520.11,145.97) ;
            \draw    (360.11,134.22) -- (360.11,145.97) ;
            \draw    (253.11,180.71) -- (356.71,180.71) ;
            \draw [shift={(359.71,180.71)}, rotate = 180] [fill={rgb, 255:red, 0; green, 0; blue, 0 }  ][line width=0.08]  [draw opacity=0] (8.93,-4.29) -- (0,0) -- (8.93,4.29) -- cycle    ;
            \draw [shift={(250.11,180.71)}, rotate = 0] [fill={rgb, 255:red, 0; green, 0; blue, 0 }  ][line width=0.08]  [draw opacity=0] (8.93,-4.29) -- (0,0) -- (8.93,4.29) -- cycle    ;
            \draw   (145.86,136.73) -- (153.58,144.7)(153.7,136.86) -- (145.74,144.57) ;
            \draw   (486.34,136.07) -- (494.06,144.04)(494.18,136.2) -- (486.22,143.91) ;
            \draw  [dash pattern={on 0.84pt off 2.51pt}]  (250.11,145.69) -- (250.11,180.71) ;
            \draw  [dash pattern={on 0.84pt off 2.51pt}]  (360.11,145.97) -- (360.11,181) ;
            \draw    (153.43,210.71) -- (487.43,210.71) ;
            \draw [shift={(490.43,210.71)}, rotate = 180] [fill={rgb, 255:red, 0; green, 0; blue, 0 }  ][line width=0.08]  [draw opacity=0] (8.93,-4.29) -- (0,0) -- (8.93,4.29) -- cycle    ;
            \draw [shift={(150.43,210.71)}, rotate = 0] [fill={rgb, 255:red, 0; green, 0; blue, 0 }  ][line width=0.08]  [draw opacity=0] (8.93,-4.29) -- (0,0) -- (8.93,4.29) -- cycle    ;
            \draw  [dash pattern={on 0.84pt off 2.51pt}]  (149.54,141.11) -- (149.54,210.71) ;
            \draw  [dash pattern={on 0.84pt off 2.51pt}]  (490.43,141.11) -- (490.43,210.71) ;
            \draw    (113.5,110.5) .. controls (144.5,54.25) and (189.5,140.75) .. (271,90.75) .. controls (352.5,40.75) and (368,97.75) .. (406.5,115.75) .. controls (445,133.75) and (486.5,42.75) .. (525.5,53.25) ;
            \draw   (171.42,136.26) .. controls (173.89,133.77) and (177.92,133.75) .. (180.41,136.23) .. controls (182.9,138.7) and (182.91,142.73) .. (180.44,145.22) .. controls (177.96,147.71) and (173.94,147.72) .. (171.45,145.25) .. controls (168.96,142.77) and (168.94,138.75) .. (171.42,136.26) -- cycle ; \draw   (171.42,136.26) -- (180.44,145.22) ; \draw   (180.41,136.23) -- (171.45,145.25) ;
            \draw   (303.7,135.69) .. controls (306.18,133.2) and (310.2,133.18) .. (312.69,135.66) .. controls (315.19,138.13) and (315.2,142.16) .. (312.72,144.65) .. controls (310.25,147.14) and (306.22,147.15) .. (303.73,144.68) .. controls (301.24,142.2) and (301.23,138.18) .. (303.7,135.69) -- cycle ; \draw   (303.7,135.69) -- (312.72,144.65) ; \draw   (312.69,135.66) -- (303.73,144.68) ;
            \draw   (443.7,135.69) .. controls (446.18,133.2) and (450.2,133.18) .. (452.69,135.66) .. controls (455.19,138.13) and (455.2,142.16) .. (452.72,144.65) .. controls (450.25,147.14) and (446.22,147.15) .. (443.73,144.68) .. controls (441.24,142.2) and (441.23,138.18) .. (443.7,135.69) -- cycle ; \draw   (443.7,135.69) -- (452.72,144.65) ; \draw   (452.69,135.66) -- (443.73,144.68) ;
            
            \draw (115.25,116.25) node [anchor=north west][inner sep=0.75pt]   [align=left] {$\beta_i$};
            \draw (244,115.75) node [anchor=north west][inner sep=0.75pt]   [align=left] {$\beta_{i+1}$};
            \draw (353.5,115.75) node [anchor=north west][inner sep=0.75pt]   [align=left] {$\beta_{i+2}$};
            \draw (512.5,116) node [anchor=north west][inner sep=0.75pt]   [align=left] {$\beta_{i+3}$};
            \draw (169.5,148.75) node [anchor=north west][inner sep=0.75pt]   [align=left] {$q_i$};
            \draw (301,146) node [anchor=north west][inner sep=0.75pt]   [align=left] {$q_{i+1}$};
            \draw (440.9,145.8) node [anchor=north west][inner sep=0.75pt]   [align=left] {$q_{i+2}$};
            \draw (282.01,188.19) node [anchor=north west][inner sep=0.75pt]   [align=left] {$dist(a,b)$};
            \draw (144.8,120) node [anchor=north west][inner sep=0.75pt]   [align=left] {$a$};
            \draw (485.05,120) node [anchor=north west][inner sep=0.75pt]   [align=left] {$b$};
            \draw (298,213.69) node [anchor=north west][inner sep=0.75pt]   [align=left] {$ED(a,b)$};

        \end{tikzpicture}

        \caption{
            Graphical proof of mindist function definition. Symbol $a$ might occur anywhere on range $\left \langle \beta_i; \beta_{i+1} \right \rangle$ and symbol $b$ on range $\left \langle \beta_{i+2}; \beta_{i+3} \right \rangle$. In this case, minimal difference is between lower bound of greater symbol and upper bound of lower symbol I.e $mindist(a,b) = \beta_{i+2} - \beta_{i+1}$. This applies for all cases ranging for symbols located directly next to each other  ($\tilde{q_i}; \tilde{q_{i+1}}$) or arbitrary symbols distant  ($\tilde{q_i}; \tilde{q_{i+n}}$). The only exception is self-symbol distance  ($\tilde{q_i}; \tilde{q_i}$) - strictly defined as 0.
        }
        \label{fig:mindist_graphical_proof}
    \end{figure}

    To make MINDIST distance working, we need to construct a lookup table for Euclidean distance between transformed symbols - internal $dist$ function. From the graphical proof (Fig. \ref{fig:mindist_graphical_proof}), it is clear that even in our implementation, $dist(q, c)$ its internal lookup $cell(q,c)$ function apply (Eq. \ref{eq:mindist_cell}). The example of such calculation for the breakpoints vector is listed in Table \ref{tab:mindist_example}.
    
    \begin{equation}
        \label{eq:mindist_cell}
        cell_{q,c} = \begin{cases}
          0 & if q = c \\
          \beta_{max(q, c)-1} - \beta_{min(q,c)} & otherwise
        \end{cases}  
    \end{equation}

    \begin{table}
        \centering
        \caption{The example of lookup table for estimated $pdf$, alphabet size $a = 6$ and breakpoints vector $\beta = [-\infty, -0.33, -0.01, 0.66, 0.97, 1.54, \infty]$. }
        \label{tab:mindist_example}
        \begingroup
            \setlength{\tabcolsep}{6pt} 
            \renewcommand{\arraystretch}{1.2}
            \begin{tabular}{l|llllll}
                       & \textbf{a} & \textbf{b} & \textbf{c} & \textbf{d} & \textbf{e} & \textbf{f}    \\ \hline
            \textbf{a} & 0.00       & 0.00       & 0.32       & 0.99       & 1.30       & 1.87 \\
            \textbf{b} & 0.00       & 0.00       & 0.00       & 0.67       & 0.98       & 1.55 \\
            \textbf{c} & 0.32       & 0.00       & 0.00       & 0.00       & 0.31       & 0.88 \\
            \textbf{d} & 0.99       & 0.67       & 0.00       & 0.00       & 0.00       & 0.57 \\
            \textbf{e} & 1.30       & 0.98       & 0.31       & 0.00       & 0.00       & 0.00 \\
            \textbf{f} & 1.87       & 1.55       & 0.88       & 0.57       & 0.00       & 0.00
            \end{tabular}
        \endgroup
    \end{table}

\section{Experiments \& Evaluation}
    We evaluated our method by means of reconstruction error and tightness of lower bound proof. As far as we know, the most suitable method to compare with is the SAX. In the next sections we describe used evaluation data sets and discuss achieved results with their implications in real life method exploitation.
    
    \subsection{Evaluation data sets description}
        \label{sect:evaluation_datasets}
        For evaluation purposes, we selected a consistent collection of 20 different time series data sets from UEA \& UCR Time Series Classification Repository \cite{bagnall_lines_keogh_ts_repository}. Table \ref{tab:evaluation_datasets}. describes a general overview for all datasets such as their time series count, train / test size, categories count and their type. The selection covers different types of time series such as Image, Sensor, Motion or Simulated; and variable lengths ranging from tens to hundreds of points with overlap into different domains such as biology, physics, chemometrics or industry in general.
        
        \begin{table}[!h]
            \centering
            \caption{Brief evaluation datasets description.}
            \label{tab:evaluation_datasets}
            \begingroup
                \setlength{\tabcolsep}{4pt} 
                \renewcommand{\arraystretch}{1.35}
                \begin{tabular}{|l|r|r|r|r|l|l|}
                    \hline
                    \textbf{Dataset} & \textbf{Train S} & \textbf{Test S} & \textbf{Avg Len} & \textbf{Classes} & \textbf{Type} & \textbf{Domain}  \\ \hline
                    Adiac            & 390                 & 391                & 176             & 37               & Image         & Biology \\ \hline
                    Beef             & 30                  & 30                 & 470             & 5                & Spectro       & Chemometrics \\ \hline
                    BeetleFly        & 20                  & 20                 & 512             & 2                & Image         & Biology \\ \hline
                    CBF              & 30                  & 900                & 128             & 3                & Simulated     & N/A \\ \hline
                    Coffee           & 28                  & 28                 & 286             & 2                & Spetro        & Chemometrics \\ \hline
                    FaceAll          & 560                 & 1690               & 131             & 14               & Image         & Surveillance \\ \hline
                    FaceFour         & 24                  & 88                 & 350             & 4                & Image         & Surveillance \\ \hline
                    Fish             & 175                 & 175                & 463             & 7                & Image         & Biology \\ \hline
                    GunPoint         & 50                  & 150                & 150             & 2                & Motion        & Surveillance \\ \hline
                    Lightning2       & 60                  & 61                 & 637             & 2                & Sensor        & Physics \\ \hline
                    Lightning7       & 70                  & 73                 & 319             & 7                & Sensor        & Physics \\ \hline
                    OSULeaf          & 200                 & 242                & 427             & 6                & Image         & Biology \\ \hline
                    OliveOil         & 30                  & 30                 & 570             & 4                & Spectro       & Chemometrics \\ \hline
                    SwedishLeaf      & 500                 & 625                & 128             & 15               & Image         & Biology \\ \hline
                    Syn. Control     & 300                 & 300                & 60              & 6                & Simulated     & N/A \\ \hline
                    Trace            & 100                 & 100                & 275             & 4                & Sensor        & Industry \\ \hline
                    TwoPatterns      & 1000                & 4000               & 128             & 4                & Simulated     & N/A \\ \hline
                    Wafer            & 1000                & 6164               & 152             & 2                & Sensor        & Industry \\ \hline
                    Worms            & 181                 & 77                 & 900             & 5                & Motion        & Biology \\ \hline
                    Yoga             & 300                 & 3000               & 326             & 2                & Image         & Surveillance \\ \hline
                \end{tabular}
            \endgroup
        \end{table}
        
\begin{figure}[]
            \centering
            \includegraphics[width=1.0\textwidth]{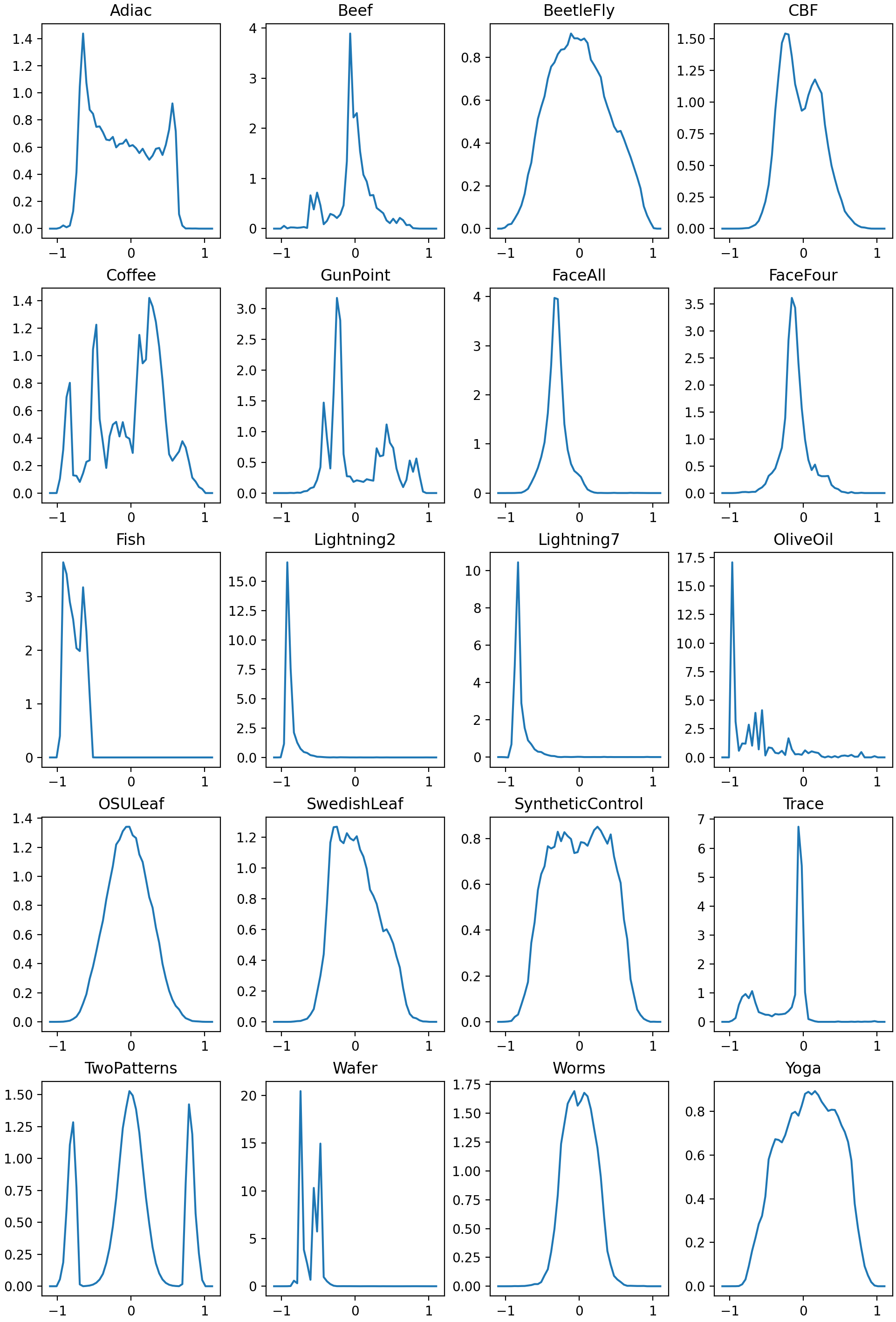}
            
            \caption{KDE distributions estimation visualizations using Epanechnikov kernel and ISJ based bandwidth depicting data distributions modality and intuition for reasoning about further evaluation tasks results.}
            \label{fig:evaluation_datasets}
        \end{figure}

        Considering our further evaluation, it is necessary to have intuition about underlying data distribution for each evaluation dataset - their shape. Figure \ref{fig:evaluation_datasets}. shows grid of all estimated data sets $pdf$s using Epanechnikov kernel and per data set calculated ISJ bandwidth parameter.  There are several multimodal datasets such as Beef, Coffee or TwoPatterns, several mixed ones with uncertain distribution like GunPoint or Trace, and clear Gaussian-like shaped datasets - BettleFly, OSULeaf or Worms.
        
    \subsection{Tightness of Lower Bound}
        Tightness of Lower Bound ($TLB$) is a key metric for method indexing performance (Eq. \ref{eq:eval_tlb}). Symbolic or in general dimensionally reduced time series representations are used for cheap database index lookups without any need to access raw timeseries stored on slow large-capacity storage. Optimal TLB ranges values from zero to one; meaning poorly or truly precise calculated symbolized time series distance compared to distance measure calculated on raw time series.
        
        \begin{equation}
            \label{eq:eval_tlb}
            TLB = \frac{MINDIST(\tilde{Q}, \tilde{C})}{D(Q,C)}
        \end{equation}  
        
\begin{figure}[!h]
            \centering
            \includegraphics[width=1.0\textwidth]{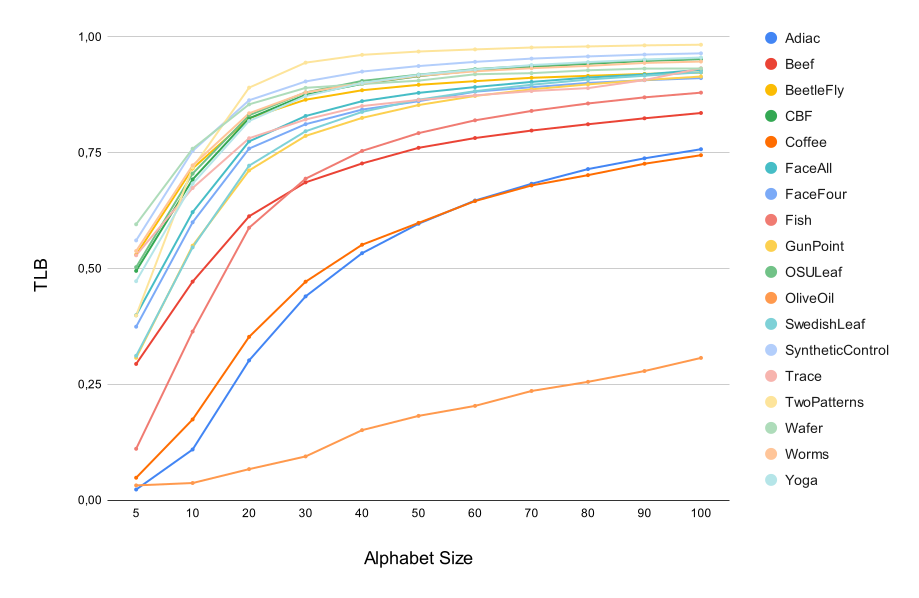}
            
            \caption{Tightness of Lower Bound performance with respect to alphabet size ranging from 5 to 100 symbols on all evaluation data sets. With increasing alphabet size, TLB is getting tighter to Euclidean distance.}
            \label{fig:evaluation_tlb}
        \end{figure}

\begin{figure}[!h]
            \centering
            \includegraphics[width=0.85\textwidth]{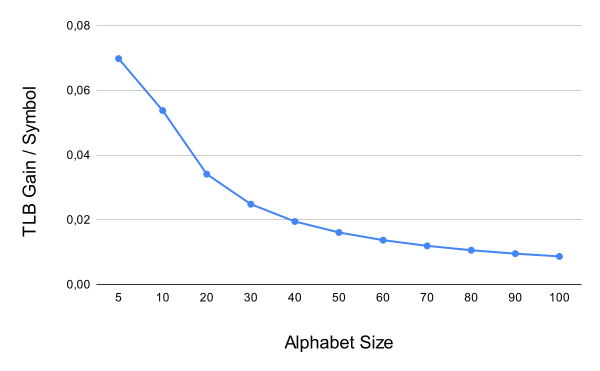}
            
            \caption{Ratio between alphabet size and TLB improvement. The more symbols we have in the alphabet, the better the results. However, a reasonable boundary seems around alphabet size 40.}
            \label{fig:evaluation_tlb_per_symbol}
        \end{figure}

        In our test scenario, we trained for each data set edwSAX model with Epanechnikov kernel, ISJ automatic bandwidth selection and PAA transformation parameter 2. After we trained our models, we evaluate TLB performance using different alphabet sizes such as $[5, 10, 20, 30, 40, 50, 60, 70, 80, 90, 100]$ symbols. For 16 out of 20 datasets we were able to reach pretty similar performance varying TLB from $0.29$ to $0.61$ for alphabet size 5 to maximum range from $0.79$ to $0.98$ for alphabet size 100. From Figure \ref{fig:evaluation_tlb}. we can see a steady relative increasing trend between alphabet size and TLB. Reasonable trade-off between alphabet size and gained TLB between 20 and 30 alphabet symbols (Fig. \ref{fig:evaluation_tlb_per_symbol}). OliveOil as an outlier in performance strongly underperforms this task reaching TLB from 0.03 for 5 symbols to only 0.31 for 100 symbols. TLB Trend is also increasing, but still underperforming the other datasets. If we look at data distribution for this dataset, we can see a strongly fragmented distribution across the whole range of values.
        
        We evaluated the relation between alphabet size and TLB precision gained by adding extra symbols to the alphabet. Having average TLB across all data sets for respective alphabet size, we see reasonable improvement of TLB per each extra symbol ranging $0.07$ per symbol, alphabet size five, to $0.02$ per symbol for alphabet size forty. Using alphabet size over 40 still improves TLB, but this improvement is not significantly comparable to less symbols alphabets.
    
    \subsection{Reconstruction Error}
        Given symbolic time series representation, it is essential to evaluate the ability to reconstruct time series back from symbolic representation. Every symbolic representation method introduces various errors by smoothing out time series details / reducing dimensionality. The lesser the reconstruction error the more details encoded in symbolic representation. Lower reconstruction error indirectly implies more time series details encoded in symbolic representation from an original time series, therefore gives more precise calculations on symbolic representations.
        
\begin{figure}[!h]
            \centering
            \includegraphics[width=1.0\textwidth]{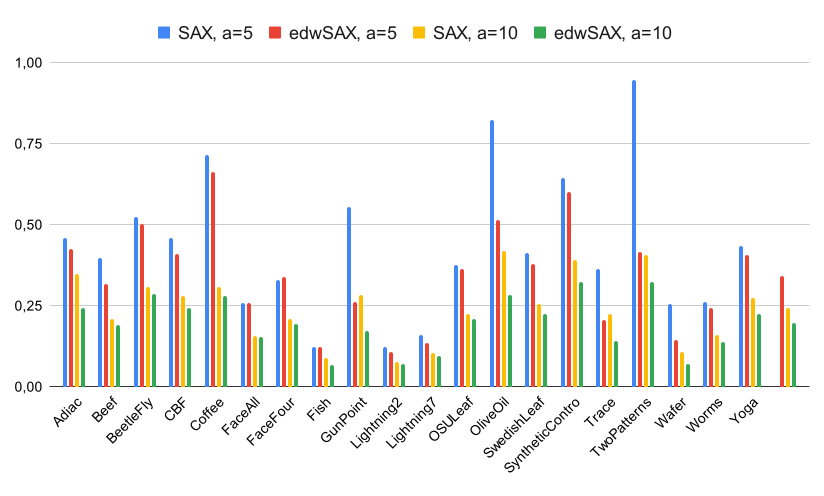}
                
            \caption{Reconstruction errors (RMSE) comparison between SAX and edwSAX with respect to alphabet size 5 and 10.}
            \label{fig:eval_reconstruction_error_overview}
        \end{figure}

        In this task, we focused on reconstruction error comparison between edwSAX and SAX methods with the respect to alphabet size evaluating RMSE metric between original and reconstructed time series. Figure \ref{fig:eval_reconstruction_error_overview} show the results comparing SAX and edwSAX for all already mentioned evaluation data sets (Chapter \ref{sect:evaluation_datasets}.) for alphabet size 5 and 10 with parameters PAA=2, Epanechnikov kernel and ISJ bandwidth selection. All mentioned results satisfy statistical significance by Wilcoxon Signed-Rank test for p=0.05. For alphabet size 5, SAX reached average RMSE reconstruction error 0.43 while edwSAX 0.34. With increasing alphabet size both methods improved their results leading SAX and edwSAX to 0.24 and 0.20 respectively. For detailed results with standard deviations, please refer to Table \ref{tab_evaluation_reconstruction_error}.
        
        \begin{table}[h]
            \centering
            \caption{Reconstruction error (RMSE) comparison between SAX and edwSAX \\ with respect to alphabet size 5 and 10.}
            \label{tab_evaluation_reconstruction_error}
            \begingroup
                \setlength{\tabcolsep}{3pt} 
                \renewcommand{\arraystretch}{1.4}
                \begin{tabular}{|l|l|l|l|l|l|l|l|l|}
                \hline
                \multirow{2}{0pt}{\textbf{Dataset}} & \multicolumn{2}{l|}{\textbf{SAX, a=5}}              & \multicolumn{2}{l|}{\textbf{edwSAX, a=5}}            & \multicolumn{2}{l|}{\textbf{SAX, a=10}}             & \multicolumn{2}{l|}{\textbf{edwSAX, a=10}}           \\ \cline{2-9} 
                                                 & \multicolumn{1}{l|}{AVG} & \multicolumn{1}{l|}{STD} & \multicolumn{1}{l|}{AVG} & \multicolumn{1}{l|}{STD} & \multicolumn{1}{l|}{AVG} & \multicolumn{1}{l|}{STD} & \multicolumn{1}{l|}{AVG} & \multicolumn{1}{l|}{STD} \\ \hline
                    Adiac & 0,46798 & 0,04336 & 0,43558 & 0,02608 & 0,35130 & 0,03192 & 0,24581 & 0,01012 \\ \hline 
                    Beef & 0,42872 & 0,05118 & 0,45018 & 0,08390 & 0,22054 & 0,02548 & 0,20472 & 0,03357 \\ \hline 
                    CBF & 0,44720 & 0,04410 & 0,40646 & 0,04595 & 0,27328 & 0,01467 & 0,24369 & 0,04257 \\ \hline 
                    Coffee & 0,72978 & 0,02463 & 0,64465 & 0,02335 & 0,30757 & 0,02409 & 0,26161 & 0,00927 \\ \hline 
                    FaceAll & 0,25646 & 0,00894 & 0,28358 & 0,01465 & 0,15228 & 0,00907 & 0,15377 & 0,00904 \\ \hline 
                    FaceFour & 0,32128 & 0,01505 & 0,34581 & 0,03310 & 0,19923 & 0,02141 & 0,19627 & 0,02353 \\ \hline 
                    Fish & 0,12055 & 0,00794 & 0,13240 & 0,01472 & 0,09163 & 0,00576 & 0,06558 & 0,00409 \\ \hline 
                    GunPoint & 0,53585 & 0,13174 & 0,27708 & 0,08271 & 0,29617 & 0,14523 & 0,14883 & 0,02309 \\ \hline 
                    Lightning2 & 0,11315 & 0,02612 & 0,12756 & 0,03903 & 0,08079 & 0,01399 & 0,08295 & 0,02091 \\ \hline 
                    Lightning7 & 0,15137 & 0,02843 & 0,17274 & 0,03469 & 0,11035 & 0,01227 & 0,10787 & 0,02266 \\ \hline 
                    OSULeaf & 0,38912 & 0,03333 & 0,39733 & 0,03857 & 0,22264 & 0,01589 & 0,21322 & 0,02027 \\ \hline 
                    OliveOil & 0,81951 & 0,00839 & 0,55603 & 0,00849 & 0,41591 & 0,00354 & 0,27051 & 0,00277 \\ \hline 
                    SwedishLeaf & 0,39812 & 0,02971 & 0,37035 & 0,03210 & 0,22593 & 0,01982 & 0,21155 & 0,02902 \\ \hline 
                    Syn. Control & 0,69695 & 0,05037 & 0,63973 & 0,05991 & 0,38357 & 0,03309 & 0,31669 & 0,02471 \\ \hline 
                    Trace & 0,37012 & 0,09105 & 0,19415 & 0,08092 & 0,28719 & 0,10500 & 0,12201 & 0,04162 \\ \hline 
                    TwoPatterns & 1,03338 & 0,09201 & 1,39209 & 0,19217 & 0,31282 & 0,06090 & 0,30089 & 0,09275 \\ \hline 
                    Wafer & 0,23574 & 0,06626 & 0,13432 & 0,05170 & 0,08131 & 0,03093 & 0,07751 & 0,03280 \\ \hline 
                    Yoga & 0,47174 & 0,03554 & 0,45632 & 0,05527 & 0,25553 & 0,01191 & 0,23796 & 0,02052 \\ \hline 
                \end{tabular}
            \endgroup
        \end{table}
        
\begin{figure}[]
            \centering
            \includegraphics[width=0.8\textwidth]{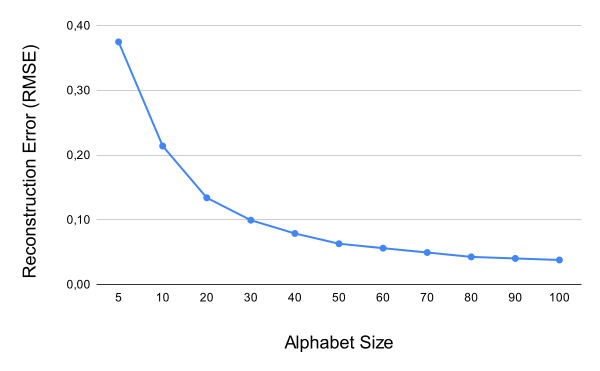}
            
            \caption{Reconstruction error (RMSE) trend across all evaluation data sets with respect to alphabet size. (edwSAX setup: Epanechnikov kernel, ISJ bandwidth selection)}
            \label{fig:evaluation_reconstruction_error_symbols}
        \end{figure}

        In the second evaluation task we compared reconstruction error with respect to alphabet size. Results (Fig. \ref{fig:evaluation_reconstruction_error_symbols}) show that with increasing alphabet size the reconstruction error is decreasing. Significant reconstruction error improvement is up to alphabet size 50, with smaller improvement for larger alphabets.
        
    To sum up achieved results, in both tasks 1) Tightness of Lower Bound and 2) Reconstruction Error edwSAX proved its potential over SAX with significantly better results on data sets that are more varying from Gaussian-like distribution. Our method reaches tightly asymptotically lower bound Euclidean distance with increasing alphabet size. This finding was also proved by average TLB per symbol improvement making our TLB increasing with every addition of symbol. Reconstruction error evaluation also shows improved performance of SAX by means of more efficient symbol representative values selection. Improved symbols selection clearly outperforms SAX on TwoPatterns dataset with strong multimodal distribution having reconstruction error 0.95 while edwSAX achieved error only 0.41.

\section{Conclusion and Future work}
    As stated in the Introduction, our main goal was to improve symbolic representation of time series with non-Gaussian data distribution. Lin et al. \cite{ref_sax_original} proposed a superior method for symbolic representation of time series - SAX. Although this approach is interesting, it does not work well for time series with non-Gaussian data distribution. We believe that we have designed an innovative solution for this problem. Our approach extends original SAX by means of dynamically captured data distribution of underlying time series and defining alternative vector of breakpoints for characters mapping. Data distribution estimation at its simplest form could be estimated through well-known and widely applicable histograms. However, this approach suffers from the ease of dynamic computation of breakpoints. An alternative solution, though with high overheads, is estimation using continuous function based estimator. Density estimators appear to be a solution to this problem. The most common variants of these estimators are kernel density estimators.
    
    This method represents a viable alternative to the original SAX method. We proved our method and output symbolic representation with a proper distance lookup table leveraging existing MINDIST distance measure lower bounds Euclidean distance. It makes our method feasible for index lookup operations. We compared our method with original SAX in two tasks: tightness of lower bound and reconstruction error. Both tasks directly/indirectly prove that our method superiors SAX with the most significant difference on multi-modal data sets.
    
    The most important limitation lies in unnecessary KDE application in case of highly Gaussian distributed data. Applying both methods in this case will result in very similar symbolic representation. KDE estimates breakpoints similar to pre-computed breakpoints from the SAX table, but with undoubtedly higher computational complexity. On the other hand, applying edwSAX without any prior knowledge of data distribution will safely produce efficient symbolic representation.
    A number of potential shortcomings need to be considered. Firstly, computational complexity of KDE and breakpoints vector recalculations needs to be considered in case of online exploitation. Secondly, the concept drift at its basis is not covered in the proposed method, though KDE with periodical recalculation is able to overcome a skew in data distribution to some extent. The third shortcoming is connected with the fact that knowledge of breakpoints vector and centroids used during discretization is crucial for further operations such as time series indexing and distance measure in general. Nevertheless, we believe that our work could be a springboard for research in the field of data distribution-aware symbolic time series representation. 
    
    This study has gone some way towards enhancing our understanding of efficient symbolic time series representation. To deepen our research we plan to design an online version of our method to tackle computational complexity with hard online processing constraints. Our results are promising and should be validated by a larger sample size time series from real-life environments.
    
\section*{Acknowledgement}
This research was supported by TAILOR, a project funded by Horizon 2020 research and innovation programme under GA no 952215 
and "Knowledge-based Approach to Intelligent Big Data Analysis" - Slovak Research and Development Agency under the contract No. APVV- 16-0213.

%
%
%
\typeout{}
\bibliographystyle{splncs04}
\bibliography{bibliography}

\end{document}